\documentclass[conference]{IEEEtran}
\IEEEoverridecommandlockouts
\usepackage{cite}
\usepackage{amsmath,amssymb,amsfonts}
\usepackage{algorithmic}
\usepackage{graphicx}
\usepackage{textcomp}
\usepackage{xcolor}

\def\BibTeX{{\rm B\kern-.05em{\sc i\kern-.025em b}\kern-.08em
    T\kern-.1667em\lower.7ex\hbox{E}\kern-.125emX}}

\begin{document}

\makeatletter
    \newcommand{\linebreakand}{%
      \end{@IEEEauthorhalign}
      \hfill\mbox{}\par
      \mbox{}\hfill\begin{@IEEEauthorhalign}
    }
    \makeatother

\title{LANet: A Lane Boundaries-Aware Approach For Robust Trajectory Prediction\\

\thanks{Accepted at the 17th IEEE International Conference on Advanced Computational Intelligence (ICACI 2025).
}
}


\author{\IEEEauthorblockN{Muhammad Atta ur Rahman}
\IEEEauthorblockA{\textit{Artificial Intelligence Creative Research Lab, ETRI} \\
\textit{University of Science and Technology, South Korea}\\
Daejeon, South Korea \\
rahman@etri.re.kr}
\and
\IEEEauthorblockN{Dooseop Choi}
\IEEEauthorblockA{\textit{Artificial Intelligence Creative Research Lab, ETRI} \\
\textit{University of Science and Technology, South Korea}\\
Daejeon, South Korea \\
d1024.choi@etri.re.kr}

\linebreakand

\IEEEauthorblockN{KyoungWook Min}
\IEEEauthorblockA{\textit{Artificial Intelligence Creative Research Lab, ETRI} \\
Daejeon, South Korea \\
kwmin92@etri.re.kr}

}
\maketitle

\begin{abstract}
   Accurate motion forecasting is critical for safe and efficient autonomous driving, enabling vehicles to predict future trajectories and make informed decisions in complex traffic scenarios. Most of the current designs of motion prediction models are based on the major representation of lane centerlines, which limits their capability to capture critical road environments and traffic rules and constraints. In this work, we propose an enhanced motion forecasting model informed by multiple vector map elements, including lane boundaries and road edges, that facilitates a richer and more complete representation of driving environments. An effective feature fusion strategy is developed to merge information in different vector map components, where the model learns holistic information on road structures and their interactions with agents. Since encoding more information about the road environment increases memory usage and is computationally expensive, we developed an effective pruning mechanism that filters the most relevant map connections to the target agent, ensuring computational efficiency while maintaining essential spatial and semantic relationships for accurate trajectory prediction. Overcoming the limitations of lane centerline-based models, our method provides a more informative and efficient representation of the driving environment and advances the state of the art for autonomous vehicle motion forecasting. We verify our approach with extensive experiments on the Argoverse 2 motion forecasting dataset, where our method maintains competitiveness on AV2 while achieving improved performance.
\end{abstract}


\section{Introduction}

\label{sec:intro}

The core challenge in predicting the future trajectory of AVs is related to enabling safe and efficient autonomous driving. It provides high accuracy in trajectory prediction, thereby making better decisions, anticipating collisions, and smoothly moving around in challenging scenarios. Recent advancements in deep and representation learning have significantly improved the performance of trajectory prediction models \cite{23, 9, 5, 32, 21} by exploiting spatial and temporal patterns in driving behaviors and road structures. However, most of them rely heavily on lane centerlines to represent road topology, which is probably insufficient for providing contextual information to make precise predictions.

 While lane centerlines capture the connectivity and directionality of lanes, which are the most critical building blocks of a vector map, there are more critical environmental constraints and traffic rules that essentially guide vehicle behavior. For example, lane boundaries offer important information on lane change prohibitions, road curvatures, and drivable areas, which are highly important for safe and rule-compliant navigation \cite{24}. Other elements, such as crosswalks, stop lines, and road edges, complement the vector map in comprehensively describing the driving scene. If these are ignored, the model cannot capture the spatial and semantic context necessary to reason for future trajectories, which can lead to suboptimal motion forecasts. We further propose that the use of various vector map features aside from lane centerlines can also improve motion forecasting performance because the driving context is represented in a richer view. By embedding different vector map features and fusion, the features for motion prediction models can possess a higher perception of semantic meanings related to the structure of roads and traffic rules. Recently, a set of studies \cite{5} investigated learning structural embeddings from vectorized maps and succeeded in boosting their trajectory prediction results.

 This study introduces an enhanced architecture of the motion forecasting model that captures multiple vector map elements beyond the lane centerline by fusing \cite{22} features to learn a comprehensive driving environment representation. Our proposed architecture enhances the trajectory prediction accuracy because richer contextual information is available. This enhances the ability of the AV to estimate the future motions of other traffic-road participants more precisely. Adding full-lane information significantly increases memory consumption; therefore, it becomes computationally expensive. In this context, we present a pruning mechanism that efficiently filters only the most salient map connections of the target agent. Thus, our model selects only the most informative lane elements and reduces the computational overhead while retaining only the critical spatial and semantic relationships pertinent to trajectory forecasting.

Our contributions are 3-fold: First, we incorporate multiple elements of the vector map, such as lane boundaries and road edges, to enrich the representation of the driving environment. This enables the model to better capture traffic rules and road constraints influencing vehicle motion. A feature fusion strategy is introduced to effectively merge the information from various vector map elements, which will allow the model to learn more holistically about the road topology and interactions of the agents. Second, since encoding more lane information increases memory consumption, we develop a learnable pruning method to filter the most useful map connections to the target agent selectively and thus reduce computation while maintaining all crucial spatial and semantic relationships to accurately model complex traffic trajectory prediction. Third, extensive experiments are conducted on the Argoverse v2 Motion Forecasting dataset to demonstrate the effectiveness of our approach, achieving superior motion prediction performance compared to existing lane centerline-based models. Our results indicate that incorporating additional vector map elements and our pruning mechanism results in more accurate and efficient motion predictions, thus being practical for real-world autonomous driving applications.

Addressing the limitations of lane centerline-based models, our approach introduces a more informative yet efficient representation of the driving environment and advances the SOTA in motion forecasting for autonomous vehicles.

\begin{figure*}
    \centering
    \includegraphics[width=\textwidth]{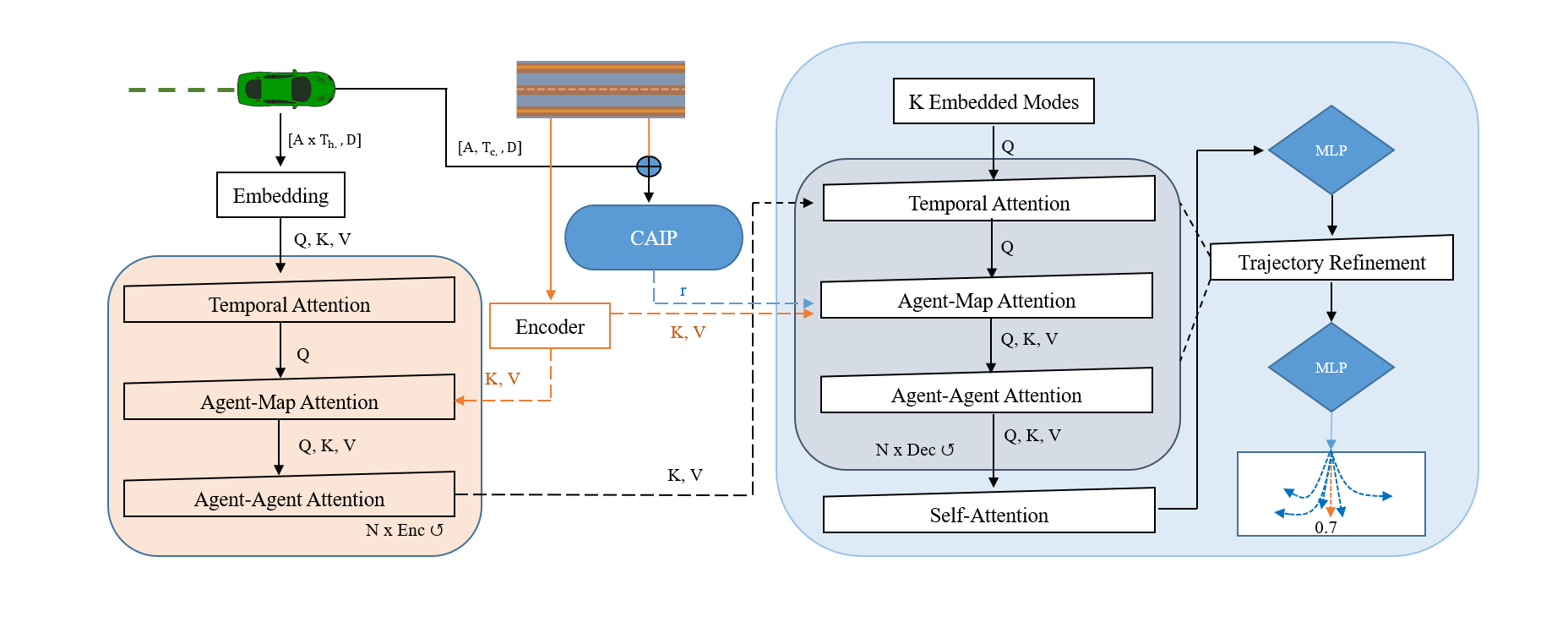}
    \caption{An overview of the proposed framework that explains modeling agent-environment interactions. Agent features are encoded and processed by temporal, agent-to-agent, and agent-to-map attention layers, which refine feature representations using relative regions, orientations, and relational embeddings. The resulting agent embeddings are sent to the decoder, which iteratively predicts and refines the multimodal trajectory.}
    \label{fig:enter-label}
\end{figure*}

\begin{figure}
    \centering
    \includegraphics[width=1\linewidth]{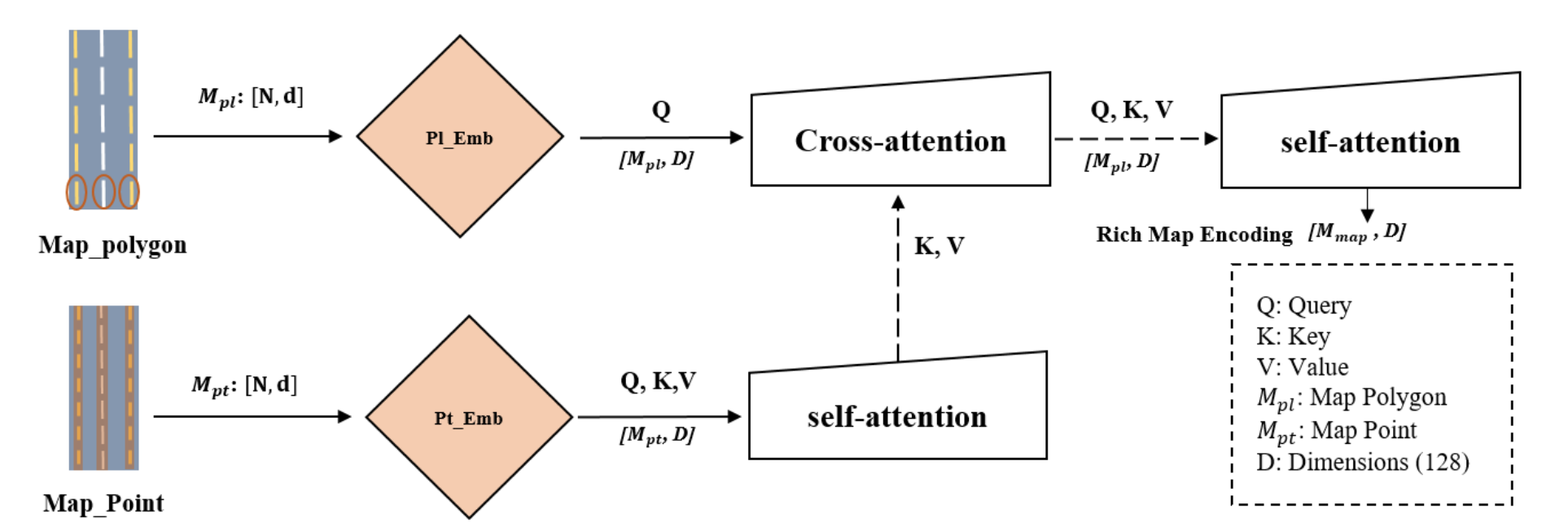}
    \caption{Overview of the Map Encoder that uses attention along with embedding layers to encode the spatial and semantic relationships between map elements. The map encoder enhanced spatial relation encoding and feature representation and allowed interaction between map elements to produce better positional encodings for richer context integration.}
    \label{fig:enter-label}
\end{figure}

\section{Related Work}

\label{sec:formatting}

\subsection{Symmetric Scene Representation}

 Scene-centric \cite{15, 16} and agent-centric \cite{5, 7, 10, 13} approaches have inherent trade-offs, sacrificing accuracy for the computational expense. To overcome these constraints, recent research \cite{9, 17, 32} has explored symmetric modeling to enhance feature fusion. HiVT \cite{9} incorporates viewpoint invariance by normalizing local context per agent and encoding relative poses directly for both local and global feature combinations. Similarly, HDGT \cite{17} and GoRela \cite{18} integrate pairwise relative positional encoding into the message-passing formulations in heterogeneous graphs. Similarly, \cite{32} encourages viewpoint invariance in spatial-temporal view by integrating time into relative positional encoding to enable continuous streaming processing. To provide a more reliable and consistent modeling method, we use a similar technique in our agent trajectory encoding framework, which enhances the model's prediction ability by better capturing dynamic interactions through the use of time-aware relative encoding.

\subsection{Context Encoding and Integration}

Driving context can be divided into two crucial categories: the temporal motion patterns of surrounding agents and the static topology of the map. Temporal networks \cite{1, 2} have been used extensively to encode trajectory data because trajectory data is sequential. Map features have been classically modeled as multi-channel bird's-eye view (BEV) images where semantic parts are encoded in various channels. These are then processed using convolutional neural networks (CNNs) to perform feature fusion and extraction \cite{3, 4}. However, rasterization results in inherent information loss and restricts the receptive field. To overcome these limitations, vectorized representations have gained popularity \cite{5, 6}, and the method of choice in recent research \cite{7, 8}. They maintain spatial consistency by encoding map features as polylines \cite{5, 7, 9, 11} or sparse graphs \cite{6, 10, 8}, from raw coordinate data. Further processing is performed through graph neural networks (GNNs) \cite{12, 13} and transformer models \cite{14}, enabling more efficient and effective feature extraction with the preservation of fine-grained spatial detail. Modern trajectory prediction models utilize transformers with factorized attention as their encoders to gain enormous efficiency gains with agent-centric hierarchically learned representations \cite{20, 15, 21} or learning the entire scene in a global coordinate system. These models remain bottlenecked by the computational expense of factorized attention. Motivated by these earlier approaches, we created our model, LANet, which uses attention mechanisms to fuse and extract spatial-temporal characteristics and encodes historical trajectories and interactions using an encoder-decoder structure.


\section{Methodology}
The methodology section begins with the problem formulation section. Next, we describe scene context encoding, context-aware interaction pruning (CAIP), and the decoder along with the training objectives.

\subsection{Problem Formulation}

In a traffic scene with $N$ agents, we aim to model the trajectory distribution $p(Y_i|X_i,C_i)$ for each agent $A_i$ based on its observed history. The positional history of $A_i$ over the past $H$ timesteps up to time $t$ is represented as $X_i=p_i(t-H:t)$. $C_i$ contains additional contextual information, including the past trajectories of surrounding agents $\{X_j\}_{j\neq i}$ and a set of $M$ map elements, including crosswalks and lane candidates $L^{(1:M)}$ available to $A_i$ at time $t$. Each lane $L_m$ is represented by $F$ equally spaced points along its centerline and boundaries. For example, at each time step \( t \), the state of the $A$-th agent includes its spatial position \( \mathbf{p}_i^t = (p_{i,x}^t, p_{i,y}^t) \), its angular orientation \( \theta_i^t \), the time step \( t \), and its velocity \( v_i^t \). Additionally, the motion vector \( \mathbf{p}_i^t - \mathbf{p}_i^{t-1} \) is also incorporated as part of the geometric attributes. The prediction module also has access to \( M \) polygons on the high-definition map, which include features such as lane centerlines, lane boundaries, and crosswalks. Each polygon is annotated with sampled points and semantic information, such as the lane type. The future positions for each agent $A_i$ for the next $T$ timesteps are given by $Y_i=p_i(t+1:t+T)$. The prediction module uses map information and agent states to estimate $K$ future trajectories for each target agent over $T$ time steps and provide a probability score for each trajectory.

\subsection{Scene Context Encoding}

\subsubsection{Map Encoding}
To represent the spatial and semantic relations among map entities (polygons and map points), we use transformers to model relational edges, evaluate structured input representations, project them to embeddings, and refine the information using recurrent message passing. A multilayer perceptron (MLP) is used to encode the raw data of the map\_polygon and map\_point. The map polygons contain the entry points of the lane centerlines, lane boundaries, and crosswalks, while the map points contain all the elements on the map. This ensures a rich representation for processing later by extracting key characteristics from both categories and continuous variables. Directed edges from points to their respective polygons were created to simulate the interactions between points and polygons. We stacked the Euclidean distance and orientation difference between the map point and the polygon, respectively, and an MLP is used to embed these spatial relationships. This enhanced representation serves as positional encoding \(r^{(pl2pt)}\) in the pl\_to\_pt attention layer and captures the spatial and orientation-based relationships between points and polygons. The same method is applied to polygon-to-polygon interactions to encode their spatial relationships and connection information.   

To encode the spatial relationships between map points, a k-nearest neighbor (KNN) graph is constructed in which each point is connected to its k-nearest points concerning the Euclidean distance, to store local spatial interactions between nearby points. This creates a set of edges that includes local spatial patterns. To maintain consistency, the relative position and angular difference between consecutive points were calculated, concatenated, and encoded using MLP to represent point-wise spatial relations and neighborhood structural links reasonably well. The network learns the spatial interdependence between points better, and such representation is the positional encoding \(r^{(pt2pt)}\) in the pt-to-pt self-attention layer where the point embedding \( x_{pt} \) is used as Q, K, and V, and given an edge set \( e_{pt \to tp} \), the attention update follows:
\begin{equation}
x_{pt} = \textit{self-att}_{{pt} \to {pt}} (x_{pt}, r^{(pt2pt)}, e_{pt \to pt})
\end{equation}
 For Pl-to-Pt interactions, polygon embeddings are influenced by updated point embeddings via cross-attention, where the polygon embedding \( x_{pl} \) acts as Q, while the point embedding \( x_pt \) serves as K and V. The edge information \( e_{pl \to pt} \) guides the attention update:
\begin{equation}
x_{pl} = \textit{cross-att}_{pl \to pt} (x_{pt}, x_{pl}, r^{(pl2pt)}, e_{pl \to pt})
\end{equation}
For Pl-to-Pl attention, a self-attention mechanism updates polygon embeddings using polygon relations. Here, \( x_{pl} \) serves as Q, K, and V, while \(r^{(pl2pl)}\) is the positional information and \( e_{pl \to pl} \) represents the global connection information between lane boundaries, lane centerlines, and crosswalks:
\begin{equation}
x_{map} = \textit{self-att}_{pl \to pl} (x_{pl}, r^{(pl2pl)}, e_{pl \to pl})
\end{equation}
allowing polygons to share spatial information. Through iterative attention updates, the model captures rich geometric and semantic relationships.

\subsubsection{Agent Features Encoding}
To encode the agent's features, the observed positional history of agents is retrieved, and displacement-based motion vectors across adjacent time steps are computed. A validity mask for agent motion information is prepared such that only pertinent history data is included. Further, the heading direction for each agent was retrieved and represented as a unit vector. The orientation and position of objects in the map are also collected for simulating agent-environment interactions. To enrich feature representation, categorical agent properties such as type are converted into embedding. A feature tensor is formed by arranging motion vector magnitudes, heading directions, and velocity magnitudes. This feature tensor is fed through an embedded layer to produce a learned representation, which is used as Query (Q), Key (K), and Value (V) in the temporal attention task. A temporal graph is constructed to offer temporal connections, and temporal edge indices are computed to note the relationships of agents. Relative positions and heading offsets based on these connections are embedded and are used as the positional encoding \(r^{(t2t)}\) in the temporal attention.
\begin{equation}
x_{\text{agent}} = \textit{temp-att}(x_a, r^{(t2t)},  e_{at \to at})
\end{equation}

where \( x_a \) represents the encoded i-th agent features, \( r^{(t2t)} \) is the positional encoding, and \( e_{at \to at} \) defines the temporal connections between each agent's historical trajectory.

For modeling agent-environment interactions, spatial edges between agents and map elements are created based on a radius-based nearest neighbor search, such that only valid connections are allowed according to the agent mask. Relative positions and orientations between agents and map elements are calculated and embedded to encode spatial dependencies and are used as the positional encoding in the agent-map attention.
\begin{equation}
x_{\text{agent}} = \textit{agent-map-att}(x_{\text{map}}, x_{\text{agent}},r^{(pl2a)}, e_{pl \to a})
\end{equation}

where \( x_{\text{map}} \) represents the encoded map features, which are used as Key (K), Value (V), and \( x_{\text{agent}} \) come from the previous attention layer (Eq 7) and are used as query (Q), \( r^{(pl2a)} \) captures the relational attributes between agents and map elements, and \( e_{pl \to a}\) defines the agent-map connections.

Agent-to-agent interactions are modeled similarly with a radius-based search for finding neighboring agents and filtering connections to only keep meaningful relationships. 

\begin{equation}
x_{\text{agent}} = \textit{agent-agent-att}(x_{\text{agent}}, r^{(a2a)}, e_{a \to a})
\end{equation}

Where \( r^{(a2a)} \) is the positional embedding, while \( e_{a \to a} \) defines the agent-agent interaction graph. The \( x_{\text{agent}} \) now contains the updated rich scene context encoding. The iterative updates refine the learned embeddings, allowing the model to learn motion patterns as well as spatial relationships. The resulting agent embeddings \( x_{\text{agent}} \) are passed to the decoder for predicting future trajectories.
\begin{figure}
    \centering
    \includegraphics[width=1\linewidth]{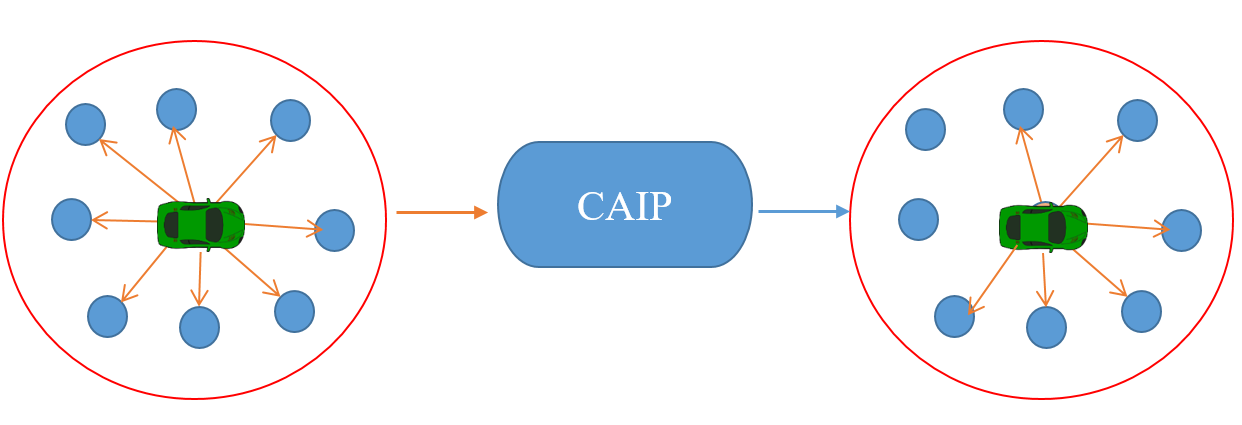}
    \caption{Overview of the Context-Aware Interaction Pruning method. First, a fully connected graph is created. Edge information in this graph is entered into the CAIP module as node features. After giving each connection an importance score, the CAIP model produces a pruned graph that only includes the most pertinent interactions.}
    \label{fig:enter-label-2}
\end{figure}
\subsection{Context-Aware Interaction Pruning}

 Quantifying the importance of spatial interactions of the target agent with the road environment is the goal of the Context-Aware Interaction Pruning (CAIP) module. To generate node features that are then introduced into the CAIP module to ascertain their significance, the relative positional and orientational differences as well as the heading alignment between the agents and map components are calculated and passed to CAIP, where, by using a feedforward neural network with multiple fully connected and several layers of ReLU activation, the CAIP is capable of catching the complex interactions between the input features and uses a sigmoid activation function to transfer them to a single scalar importance score. The network architecture consists of multiple fully connected layers with ReLU activations. 
\begin{equation}
s = \sigma \left( W \cdot \textit{ReLU} \left( W x + b \right) + b \right)
\end{equation}
 Where \( x \) represents the input node feature, \( W \) is the weight, \( b \) is the bias, and \( s \) is the final output termed the importance score after applying the sigmoid activation function. A learnable threshold parameter was incorporated, allowing the model to adaptively prune less important interactions during training. To remove irrelevant interactions and clean the edge index and its associated relational properties, the module generated a binary valid mask. These cleaned interactions are also altered by the trick thresholding function, which applies a systematic weighting scheme based on a calculated score matrix. Only the most important relationships are subject to additional calculations owing to the thresholding trick mechanism, which streamlines the selection process by normalizing scores using a temperature-scaled sigmoid function. 
\begin{equation}
v_{\text{pruned}} = v \cdot \frac{\sigma \left( -\frac{S - \theta}{\tau} \right)}{\sum_{j} \sigma \left( -\frac{S_j - \theta}{\tau} \right)}
\end{equation}
Where \( \sigma(\cdot) \) is the sigmoid function, \( S \) is the score matrix, \( \theta \) is the learnable threshold, \( \tau \) is the temperature parameter, and \( v_{\text{pruned}} \) is the final pruned feature representation. By doing this, the learned representations for downstream trajectory prediction are of higher quality, and only the most pertinent spatial relationships are retained.

\subsection{The Decoder}

 The decoder uses a DETR-like \cite{19} structure to predict future trajectories by predicting and refining the encoded agent features with several attention mechanisms and aggregating the scene context. It begins by retrieving the most recently observed agent position and heading and converting the heading to a unit vector representation. The encoded agent features \(x_{\text{agent}}\) are duplicated across several trajectory modes, and mode embeddings are added to encode the multimodal motion hypotheses. The embedded modes are used as Query (Q), and the encoded agent features are used as Key (K) and Value (V) in the temporal attention. To establish correspondence between past states and future predictions, a temporal interaction graph is constructed, mapping valid past states to prediction steps. The relative spatial-temporal position and heading are computed, embedded,  and used as positional embedding \(r\) in the temporal attention. The attention mechanism follows the same structure that is used in the encoder. To keep trajectories under environmental constraints, agent-map attention identifies relevant map features via radius-based nearest neighbor searching and learns to detect spatial relations. The edge information is sent to CAIP, which returns the most important connections and their corresponding relations to minimize computational overhead and concentrate on the most influential connections. Such relations are then encoded and used as positional information \(r\) in the agent-map attention process, where the map embeddings are used as Key (K) and Value (V), and the updated modes are used as Query (Q). Agent-agent attention models interactions at the most recent time step by forming a spatial graph among agents based on nearness and validity. This allows the modes to include motion dependency and potential interaction, enabling effective and realistic multi-agent interactions to form anchor-free trajectory modes. The decoder recursively revises these anchor-free proposals by iteratively updating mode embeddings through multiple attention layers, followed by another self-attention to enhance diversity in anchor-free predicted modes. \\
 In the final refinement step, additional attention layers that follow the same structure as the proposal section further refine the anchor-free trajectory modes, and a mode probability is computed to rank the modes' probabilities. A confidence score representing the confidence of the model is assigned to every heading prediction. The proposed and refined positions, heading angles, and confidence scores for multiple trajectory modes are finally generated by the decoder, enabling rich and precise trajectory prediction given past motion, scene constraints, and multi-agent interactions.

\subsection{Training Objectives}
We use a mixture of Laplace distributions to model the future trajectory of the i-th agent, as described in \cite{8, 32}.

\begin{equation}
  g\left( \{ q_{s_j} \}_{s=1}^{S'} \right) = \sum_{m=1}^{M} \alpha_{j,m} \prod_{s=1}^{S'} \textit{Laplace} \left( q_{s_j} \mid \nu_{s_j,m}, \beta_{s_j,m} \right)
\end{equation}
Where \( \{ \alpha_{j,m} \}_{m=1}^{M} \) represents the mixing coefficients, where the \( m \)-th mixture component’s Laplace density at the time step \( s \) is defined by the location \( \nu_{s_j,m} \) and the scale \( \beta_{s_j,m} \). A classification loss \( L_{\text{cls}} \) is employed to optimize the mixing coefficients predicted by the refinement module in the decoder to minimize the negative log-likelihood as described in Equation 9. The locations' gradients are topped and scaled to maximize the mixing coefficients exclusively. The winner-take-all technique \cite{25} is used to optimize the positions and sizes of the decoder modules, involving backpropagating only the best-predicted trajectory.  The final loss function combines the trajectory proposal loss \( L_{\text{dec-propose}} \), refinement loss \( L_{\text{dec-refine}} \), and classification loss \( L_{\text{cls}} \) for end-to-end training:
\begin{equation}  
L = L_{\text{dec-propose}} + L_{\text{dec-refine}} + \lambda L_{\text{cls}}
\end{equation}

Where \( \lambda \) is used to balance regression and classification.

\begin{table*}[ht]
\caption{Results of various methods on trajectory prediction metrics from the Argoverse 2 motion forecasting test leaderboard. Baselines that have employed ensembling are marked with the symbol "*".}
\centering
\begin{tabular}{|c|c|c|c|c|}
\hline
\textbf{Method} & \textbf{b-minFDE$_6$} $\downarrow$ & \textbf{minADE$_6$} $\downarrow$ & \textbf{minFDE$_6$} $\downarrow$ & \textbf{MR$_6$} $\downarrow$ \\ \hline
THOMAS \cite{28}  & 2.16 & 0.88 & 1.51 & 0.20 \\ \hline
GoRela \cite{18} & 2.01 & 0.76 & 1.48 & 0.22 \\ \hline
MTR \cite{11} & 1.98 & 0.73 & 1.44 & 0.15 \\ \hline
GANet \cite{29} & 1.96 & 0.72 & 1.34 & 0.17 \\ \hline
QML$^*$ \cite{31} & 1.95 & 0.69 & 1.39 & 0.19 \\ \hline
BANet \cite{24}& 2.03 & 0.73 & 1.38 & 0.18 \\ \hline
BANet$^*$ \cite{24} & 1.92 & 0.71 & 1.36 & 0.19 \\ \hline
QCNet \cite{32} & 1.907 & 0.651 & 1.290 & 0.164 \\ \hline
\textbf{Ours}  & \textbf{1.900} & \textbf{0.647} & \textbf{1.276} & \textbf{0.160} \\ \hline
\end{tabular}
\label{tab:results}
\end{table*}

\section{Experimental Results}

We first go over our experimental setup, including datasets and evaluation metrics, and to determine how each unique design contributes to our network, a range of ablation studies is carried out. We report extensive comparisons with the SOTA methods on the Argoverse v2 motion forecasting benchmark. Along with this, qualitative results and the conclusion are included in this section.

\begin{table*}[ht]
\caption{Ablation study on CAIP features in LANet models. The results are from the validation set of Argoverse 2. The \textbf{bold} indicates the best results.}
\centering
\renewcommand{\arraystretch}{1.5} 
\setlength{\tabcolsep}{8pt} 
\begin{tabular}{|c|cccc|c|c|c|c|}
\hline
\textbf{Model} & \multicolumn{4}{c|}{\textbf{CAIP Threshold ($\theta$)}} & \textbf{b\_minFDE\textsubscript{6}} & \textbf{minADE\textsubscript{6}} & \textbf{minFDE\textsubscript{6}} & \textbf{MR\textsubscript{6}} \\
\cline{2-5}
              & 0.5 & 0.6 & \textbf{0.7} & 0.8 &  &  &  & \\
\hline
LANet-1 & $\checkmark$ & - & - & - & 1.936 & 0.737 & 1.300 & 0.174 \\ 
\hline
LANet-2 & - & $\checkmark$ & - & - & 1.911 & 0.732 & 1.290 & 0.167 \\ 
\hline
\textbf{LANet-3} & - & - & $\checkmark$ & - & \textbf{1.887} & \textbf{0.725} & \textbf{1.265} & \textbf{0.162} \\ 
\hline
LANet-4 & - & - & - & $\checkmark$ & 1.905 & 0.730 & 1.288 & 0.166 \\ 
\hline
\end{tabular}
\label{tab:ablation_lanet}
\end{table*}

\subsection{Dataset}

We validated our method on the Argoverse 2 Motion Forecasting Dataset \cite{26}, a large-scale autonomous driving dataset specifically created for motion forecasting and trajectory prediction. The AV2 is an expanded version of the original Argoverse dataset \cite{27} with richer scenery, more sensor data, and higher-quality annotations. It comprises 250K sequences from 6 cities in urban areas, featuring dense interactions of ten object categories, including vehicles, buses, bicycles, motorcyclists, and pedestrians.

Argoverse 2 has one of the most notable enhancements with its 5-second history trajectory and 6-second prediction horizon, capable of more accurate long-term motion forecasting. HD maps with traffic control, lane geometry, and connectivity data provide accurate scene understanding support. The scenarios are further split into 199908 for training, 24988 for validation, and 24984 for testing, giving a well-organized benchmark to compare the models. With more realistic agents and challenging real-world scenarios to simulate, Argoverse 2 helps advance research in autonomous driving, safety features, and smart transportation planning.

\subsection{Evaluation Metrics}
We adopt the official evaluation metrics for the AV2 baseline to benchmark motion forecasting model performance: minADE, minFDE, brier-minFDE, and MR. 

\begin{itemize}
    \item \textbf{minFDE}: Stands for Minimum Final Displacement Error, the L2 gap between the endpoint of the ground truth and the best-predicted trajectory. The trajectory with the lowest endpoint error is referred to as the best in this context.
    \item \textbf{MinADE}: Stands for Minimum Average Displacement Error. The average L2 difference between the ground truth and the best-predicted trajectory. The trajectory with the lowest endpoint error is referred to as the best in this context.
    \item \textbf{Miss Rate (MR)}: Based on endpoint error, the number of cases in which none of the predicted trajectories fall within 2.0 meters of the ground truth.
    \item \textbf{B-minFDE}: Minimum final displacement error is comparable to Brier-Minimum final displacement error. Here, the distance between the endpoint L2 is simply increased by \( (1.0 - p)^2 \), where \( p \) is the likelihood of the best-predicted trajectory.
\end{itemize}

\subsection{Quantitative Results Comparison}
Table 1 provides a comparative assessment of various trajectory prediction baselines to four main evaluation metrics, which provide a quantitative assessment of the accuracy of predicted trajectories. Our approach is the lowest on all measures of evaluation, further proving its superiority against existing methods. This highlights its capacity to make more accurate trajectory predictions by effective modeling of complex road geometry and vehicle dynamics. On the other hand, methods such as THOMAS exhibit comparatively higher errors in all categories, indicating a lower capacity to accurately predict future trajectories. Other baselines, such as GoRela, MTR, GANet, and QCNet, exhibit competitive performance. Notably, our method surpasses BANet, which also employs lane boundary information, in demonstrating the superiority of using environmental context for refining trajectories. In addition, the results verify LANet as a well-performing approach, demonstrating its ability to integrate road geometry and dynamic interactions to improve motion forecasting. LANet's high performance on the test set is a testament to how effective it is in understanding road structures, interpreting intricate driving situations, and producing accurate and dependable predictions for trajectories. This observation nicely underscores the critical role of structured scene understanding to further enhance trajectory prediction models with additional context-aware motion forecasting in autonomous driving.

\begin{figure*}
    \centering
    \includegraphics[width=\textwidth]{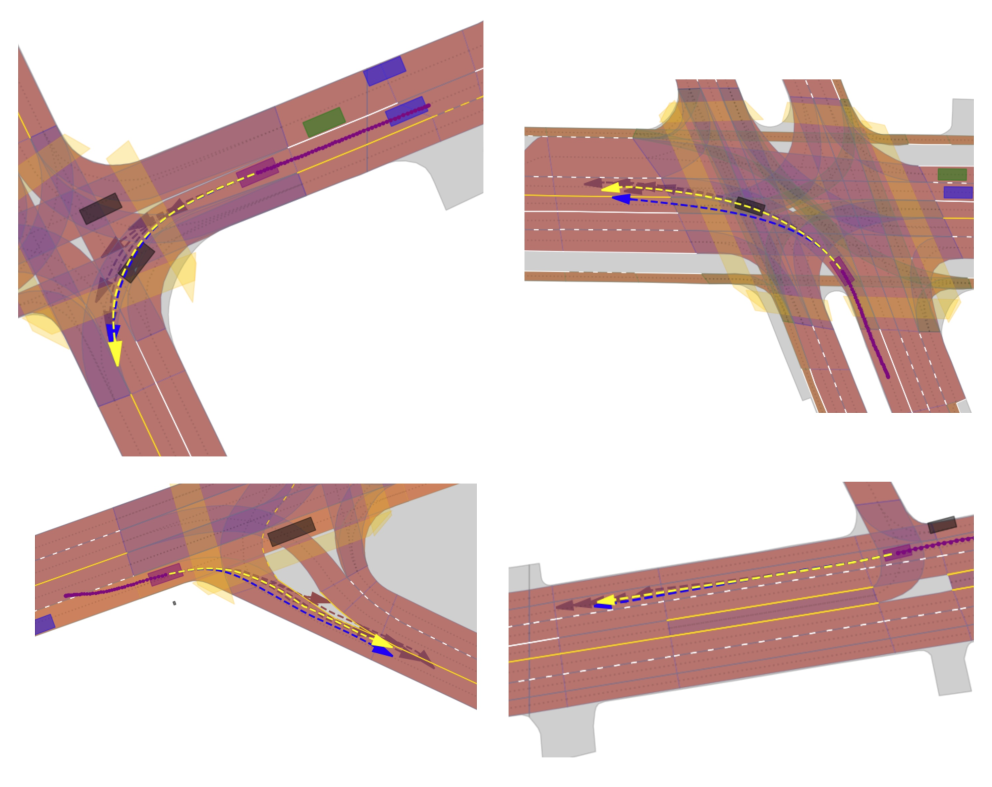}
    \caption{Qualitative results on the Argoverse 2 validation set. The target agents’ bounding boxes and ground-truth trajectories are shown in \textbf{purple} and \textbf{blue}, the models’ predictions are shown in \textbf{brown}, and the best proposal is shown in \textbf{yellow}.}
    \label{fig:qualitative-results}
\end{figure*}

\subsection{Ablation Study}

Table 2 shows the influence of varying CAIP threshold values on LANet models and the comparison of varying threshold values. The baseline model with the lowest threshold has mediocre accuracy in predicting trajectories. The prediction accuracy improves marginally, and missed predictions reduce with increased thresholds, indicating improvement in trajectory modeling. Maximum performance is achieved at the threshold (0.7) used, where the model yields the smallest prediction error and most reliable results. Raising the threshold beyond this point only increases computational cost and reduces performance because noisy links are being incorporated into the analysis. These findings indicate that it is important to employ an optimal threshold to achieve maximum accuracy and reduce inefficiency when predicting trajectories.

\subsection{Qualitative Results}
Figure 4 depicts our trajectory prediction model based on the Argoverse 2 validation set's qualitative results. The visual displays four different urban traffic scenarios that include intersections, a merging lane, and a multi-lane urban scene. The best proposal is highlighted in yellow, and ground-truth trajectories (blue) are contrasted with model-predicted trajectories (brown). The generally good agreement between predictions and actual vehicle trajectories demonstrates the model's accuracy, mainly in longer-term prediction. However, minor alterations in complicated circumstances, such as intersections and merging lanes, reveal issues with uncertainty management and nonlinear road networks. The findings show that, while models are useful for forecasting vehicle behavior, further work is needed to improve performance in dynamic urban environments.

\subsection{Conclusion}

In conclusion, this study significantly boosts motion forecasting for autonomous vehicles by leveraging a richer and more informative representation of the driving world. By incorporating multiple vector map features, such as lane boundaries and road age, we enable the model to select important traffic rules and road constraints that influence vehicle behavior. Introducing the feature fusion method enhances the model's ability to learn road topology and agent interactions holistically, and the proposed learnable pruning mechanism successfully prunes the computational overhead by selecting only the most salient map connections. Experiments on the Argoverse v2 Motion Forecasting dataset demonstrate that our approach outperforms SOTA lane-centerline-based models in motion forecasting. This study not only addresses the limitations of current models but also opens the door to more stable, rule-compliant, and efficient autonomous driving systems, with promising opportunities for real-world applications.

\subsection{Acknowledgement}
This work was supported by the Institute of Information \& Communications Technology Planning \& Evaluation (IITP) grant funded by the Korean government (MSIT) (No. RS-2024-00341055, Development of reinforcement learning-based automated driving AI software technology for optimal driving behavior decision in hazardous situations on congested roads).

{\small
\bibliographystyle{ieee_fullname}
\bibliography{egbib}
}

\end{document}